\documentclass[cameraready]{Interspeech}

\usepackage{xcolor}
\usepackage{multirow}
\usepackage{makecell}
\usepackage{amsmath}
\usepackage{mathtools}
\usepackage{subcaption}
\usepackage{stfloats}
\usepackage{tablefootnote}
\usepackage{hyperref}

\title{NaturalFlow: Reducing Disruptive Pauses for Natural Speech Flow in Simultaneous Speech-to-Speech Translation}

\author[affiliation={1}, equalcontribution]{Dongwook}{Lee}
\author[affiliation={1}, equalcontribution,]{Youngho}{Cho}
\author[affiliation={2}]{Sangkwon}{Park}
\author[affiliation={3}, daggernote]{Heeseung}{Kim$^{\dagger}$}
\author[affiliation={1,2}, daggernote]{Sungroh}{Yoon$^{\dagger}$}

\address{
    $^1$IPAI and $^2$ECE, Seoul National University, Seoul, 08826, Korea\\
    $^3$Department of AI, University of Seoul, Seoul, 02504, Korea
}

\email{\{dwsmart32, youngh36, tkdrnjs0621, sryoon\}@snu.ac.kr \quad gmltmd789@uos.ac.kr} 

\usepackage{comment}

\begin{document}

\maketitle

\begin{abstract}
    Simultaneous speech-to-speech translation aims to enable near-real-time communication by minimizing latency, offering a compelling, real-time alternative to the high latency of consecutive translation. However, the excessive pursuit of low latency often results in fragmented chunk-wise speech. Consequently, listeners are subjected to an unnatural acoustic flow punctuated by frequent pauses, which could increase their cognitive load. To bridge this gap, we introduce a fluency-aware optimization framework designed to discover the sweet spot between the low-latency benefits of simultaneous translation and the natural flow of consecutive translation. Our framework minimizes inter-chunk silences by leveraging model-internal signals, including linguistic diversity and induced temporal variability in speech durations. Experiments on short- and long-form benchmarks show that our framework produces natural speech flow while maintaining competitive latency and translation quality.
\end{abstract}

\textbf{Demo:} \href{https://naturalflows2st.github.io/naturalflow/}{https://naturalflows2st.github.io/naturalflow/}

\section{Introduction}

Speech translation is commonly categorized into two paradigms: consecutive and simultaneous. Consecutive translation generates the target speech only after a complete utterance has been received, ensuring high translation fidelity and a natural, continuous acoustic flow at the expense of significant latency. In contrast, simultaneous translation drastically reduces this delay but introduces profound cognitive challenges. As described by the ``tightrope" hypothesis~\cite{Gile1999}, human simultaneous interpreters must continuously coordinate listening, memorization, and speech production. To manage this substantial cognitive load and control latency, interpreters are forced to segment the input and deliver information in short, disjointed bursts, leading to an inherently fragmented acoustic delivery~\cite{Gile1999, Garcia2020}.

Inspired by these challenges, significant research efforts have been dedicated to advancing machine simultaneous speech-to-speech translation (Simul-S2ST). Recent breakthroughs, ranging from cascaded architectures supporting streaming ASR and decoding processes~\cite{direct_simul_chen_2021} to end-to-end models~\cite{seamless_communication_2023, streamspeech_zhang_2024, hibiki_labiausse_2025}, have considerably narrowed the performance gap between consecutive and simultaneous translation systems. However, in pursuit of minimal delay, current speech-to-speech translation (S2ST) models often generate fragmented outputs, similar to human interpreters. By relying on rigid, chunk-wise processing policies~\cite{stacl_ma_2019} that release partial content as soon as sufficient semantic context becomes available, these systems generate speech plagued by frequent pauses and unnatural acoustic flow.

This lack of fluency has significant consequences for end users. Prior studies in simultaneous interpreting~\cite{Rennert2010} demonstrate that fragmented speech substantially degrades human judgments of translation quality. Frequent pauses and disfluencies negatively impact listeners' perceptions of accuracy and overall intelligibility, even when the informational content is preserved~\cite{Rennert2010, christodoulides2014prosodic}. Despite fluency being a critical component of user experience, current S2ST research has predominantly focused on optimizing the traditional quality-latency trade-off, such as balancing BLEU scores against lag-related metrics, leaving the optimization of pause-driven acoustic flow comparatively underexplored~\cite{seamless_communication_2023, stacl_ma_2019}.

\begin{figure*}[t]
  \centering
  \includegraphics[width=\textwidth]{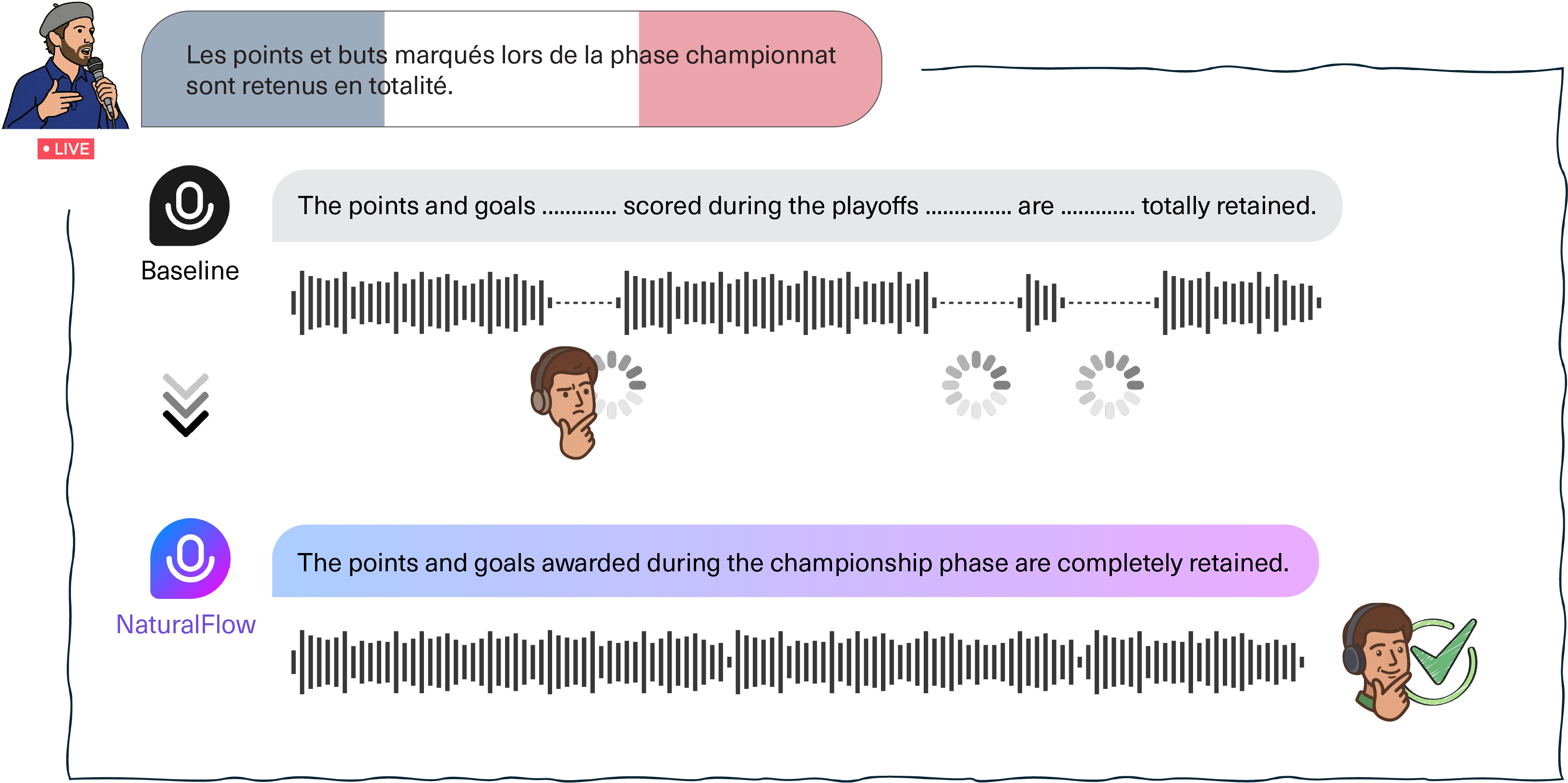}
    \caption{Comparison of translation outputs on a real example from the CVSS-C test set. Our model produces a natural flow with fewer pauses compared to the baseline.}
  \label{fig:translation_figure}
\end{figure*}

To address this challenge, we introduce a fluency-aware optimization framework designed to minimize unnatural pauses between translated chunks by leveraging the inherent generative flexibility of large language models (LLMs). Because LLM-based translation models estimate probability distributions over a vast vocabulary, they can express a single source concept through multiple linguistically valid paraphrases of varying lengths. As illustrated in Figure~\ref{fig:translation_figure}, a baseline system translates a segment concisely---resulting in an output like, ``The points and goals (pause) scored during the playoffs (pause) are totally retained"---where frequent pauses are required to wait for incoming source context. In contrast, by generating a lexically different but semantically equivalent phrase such as ``awarded during the championship phase'', the model can lengthen the duration of the spoken output by selecting paraphrases with more syllables or longer articulation time. This extended articulation gives the system additional time to continuously ingest the incoming source audio without halting the acoustic delivery. Consequently, the model circumvents the need for disruptive pauses, fostering a seamless, continuous flow.

However, optimizing this behavior is not naturally formulated as a standard supervised learning problem. For a given source segment, there is often no single gold translation that uniquely represents the optimal balance between semantic fidelity and acoustic fluency. Instead, several candidate translations may preserve the intended meaning, yet differ in their ability to lengthen spoken delivery and avoid pause insertion. We therefore formulate the problem as preference-based learning, where pairwise judgments between candidate outputs provide direct supervision for the desired trade-off between translation adequacy and continuous acoustic flow.

To this end, we introduce NaturalFlow, a speech-to-speech translation model trained to generate more continuous and natural acoustic delivery while preserving translation quality. Built on top of the Hibiki model~\cite{hibiki_labiausse_2025}, NaturalFlow is trained with Direct Preference Optimization (DPO) using a novel preference data construction methodology called Silver-Medal Preference, which jointly optimizes two potentially conflicting objectives: minimizing silence ratio and preserving translation fidelity.

We validate our framework across four benchmarks: CVSS-C~\cite{cvss_jia_2022}, VoxPopuli~\cite{voxpopuli_wang_2021}, mTEDx~\cite{mtedx_salesky_2021}, and Audio-NTREX~\cite{hibiki_zero_labiausse_2026}, covering various domains and utterance lengths. Our experiments demonstrate that the proposed method reduces the silence ratio while preserving translation quality and other latency-related metrics. Ultimately, human evaluation confirms that our S2ST model generates continuous, natural-sounding translations that are preferred by listeners over baseline systems.

\section{Related Work}

\subsection{Fluency in interpreting: pauses and perceived quality}
Fluency is a central criterion in interpreting quality assessment, but it has been operationalized through a heterogeneous set of temporal and disfluency-related correlates rather than a single agreed-upon construct~\cite{Mead2005}. A common thread across this literature is that fluency is strongly tied to real-time production under processing constraints: interpreters must listen, retain information, and speak simultaneously, making breakdown phenomena—pauses, hesitations, repairs, and repetitions—salient indicators of performance under load~\cite{Gile1999}. In this sense, fluency is often discussed as smooth and continuous delivery under time pressure, aligning with broader spoken-language accounts that define fluency as rapid and efficient online formulation and articulation~\cite{Lennon2000}.

A substantial body of work has therefore focused on how pauses manifest in simultaneous interpreting and how they relate to perceived fluency. Methodologically, pauses are typically quantified via duration- and frequency-based measures (e.g., number of silent pauses, mean pause duration), as well as composite temporal ratios such as phonation time ratio (PTR), which captures the proportion of speaking time relative to total time~\cite{Mead2005}. Empirically, descriptive analyses show that silent pauses and other disfluencies are pervasive in interpreted output and reflect the interpreter’s coordination of concurrent demands~\cite{Tissi2000}. Beyond description, perceptual work demonstrates that fluency-related cues can substantially influence subjective quality judgments. In particular, controlled studies report that manipulating fluency (including pause patterns) changes listeners’ ratings of interpreting quality, and can negatively bias perceived accuracy and intelligibility even when informational content is held constant~\cite{Rennert2010}. Complementary evidence from (para)linguistic correlates of perceived fluency further supports that pause behavior and temporal delivery features are robust predictors of fluency impressions in simultaneous interpreting~\cite{Han2015}. Taken together, interpreting research highlights pauses as a significant factor in perceived fluency and user experience—suggesting that, for simultaneous S2ST, minimizing disruptive inter-chunk silences is a key consideration rather than a secondary aesthetic concern.

\subsection{Speech-to-speech translation}
Early speech translation systems were predominantly \emph{cascaded} systems comprising automatic speech recognition (ASR), machine translation (MT), and text-to-speech (TTS) modules, which can incur error propagation and latency through intermediate decoding and re-synthesis. Motivated by these limitations, end-to-end speech-to-text translation (S2TT) emerged as a direct mapping from source speech to target text~\cite{berard2016,weiss2017}.

Speech-to-speech translation (S2ST) extends this line by generating target speech, either via an S2TT+TTS cascade or with \emph{direct} S2ST models. Representative directions include end-to-end speech-to-speech modeling~\cite{jia2019}, discrete-unit-based S2ST that predicts self-supervised units and can optionally emit both text and speech~\cite{lee2022}, and unified foundation models covering ASR/S2TT/S2ST at scale~\cite{seamlessM4T2023}. For simultaneous S2ST, the additional key challenge is deciding \emph{when} to speak while the source audio unfolds. Recent systems address this with multilingual streaming model families~\cite{seamless_communication_2023}, multi-task frameworks that jointly learn translation and a simultaneous policy~\cite{streamspeech_zhang_2024}, or decoder-only multi-stream formulations that jointly produce text and audio tokens for high-fidelity simultaneous speech translation~\cite{hibiki_labiausse_2025,hibiki_zero_labiausse_2026}.

Despite this rapid progress, much of the simultaneous S2ST literature still frames progress primarily through the \emph{quality--latency} trade-off (e.g., translation metrics versus onset delay and lagging behavior) and treats the resulting pause patterns as a secondary byproduct of chunking and policy decisions. However, as highlighted by the interpreting literature discussed above, pause-driven delivery can substantially shape perceived quality and intelligibility even when semantic content is comparable. In this work, we therefore shift the objective to explicitly improving \emph{acoustic continuity}, by directly targeting disruptive inter-chunk silences while preserving translation quality and latency.

\subsection{Optimization with translation models: preference learning for real-time behavior}
Preference-based optimization provides a practical alternative to RL-style alignment for shaping generation behavior from pairwise preferences. Direct Preference Optimization (DPO) turns alignment into a stable supervised objective over preferred and dispreferred outputs, avoiding explicit reward modeling and online RL~\cite{DPO2023}. This is attractive for translation, where multiple candidates can be sampled for each input and compared through relative preferences.

In MT, preference optimization has been used to directly improve model outputs via carefully constructed preference pairs (e.g., contrastive preferences that discourage near-miss translations)~\cite{xu_cpo_2024}. For \emph{simultaneous} MT, preference and policy optimization is especially relevant because desired behavior is time-dependent: systems must decide \emph{when} to write as the source unfolds. SimulPL shows that human preferences can be incorporated to improve streaming behavior under latency constraints~\cite{SimulPL2025}, and SeqPO-SiMT further optimizes sequential policies for better quality--latency trade-offs in multi-step streaming settings~\cite{seqpo_xu_2025}.
However, existing preference-optimized streaming systems still primarily target quality--latency trade-offs and policy efficiency. In contrast, we encode preferences to directly penalize disruptive \emph{silence} in simultaneous S2ST while constraining translation quality and latency, aligning the model toward more continuous acoustic delivery.

\section{Preliminary}
\subsection{Simultaneous S2ST model}

To achieve this near-instantaneous communication, end-to-end S2ST systems must be capable of processing incoming audio streams while synchronously generating translations. In this work, we adopt Hibiki~\cite{hibiki_labiausse_2025} as our baseline architecture and apply our proposed optimization framework to it. 

Hibiki is designed to process source speech while synchronously generating target speech and text. The raw audio is first discretized using Mimi~\cite{défossez2024moshispeechtextfoundationmodel}, a neural audio codec operating at a frame rate of $f_r = 12.5 \text{ Hz}$ with $Q=16$ codebook levels. This allows the model to capture coarse semantic meaning as well as fine acoustic details such as speaker timbre. It takes the source audio stream as input and jointly predicts target audio tokens and a word-level aligned text stream, where the text tokens are padded to match the length of the corresponding audio tokens so that both sequences are temporally aligned at the same resolution. Rather than relying on external control policies, it implicitly learns how long to accumulate additional source context and when to emit a translation. This behavior is enabled by temporally aligned training data constructed through a weakly supervised procedure: as the source context is gradually expanded, the model tracks the log-likelihood of the next target token and identifies alignment points where a sharp increase indicates that sufficient contextual information has been accumulated for confident prediction.

Although this alignment strategy reflects the spirit of simultaneous interpretation, it is also the main culprit behind the high silence ratio in the generated speech output. Since the model is trained to emit speech right after these contextual spikes from the translation LLM, it results in a fragmented acoustic flow that places a high cognitive burden on listeners. To address this limitation, we introduce a fluency-aware preference optimization framework that reduces excessive inter-chunk silences while preserving translation quality and latency using an AI-driven preference dataset.

\subsection{Direct preference optimization}
Aligning generative models with desired behaviors traditionally relies on reinforcement learning from human or AI feedback (RLHF/RLAIF)~\cite{lee2024rlaifvsrlhfscaling, ouyang2022traininglanguagemodelsfollow}, which requires training an external reward model and applying complex optimization algorithms like PPO~\cite{schulman2017proximalpolicyoptimizationalgorithms}. To streamline this pipeline, Direct Preference Optimization (DPO)~\cite{DPO2023} eliminates the need for an explicit reward model by directly optimizing the policy over a static preference dataset. This bypass of intermediate reward modeling makes DPO naturally well-suited for large-scale, offline AI-generated feedback.

Given a source speech input $x$, let $y_c$ denote the preferred (chosen) target speech response and $y_r$ denote the less preferred (rejected) response. Our offline preference dataset is constructed as $\mathcal{D} = \{(x, y_c, y_r)\}$. The standard DPO objective updates the policy $\pi_{\theta}$ relative to a reference policy $\pi_{\mathrm{ref}}$ by minimizing the following loss:

\begin{equation}
\label{eq:dpo}
\resizebox{0.9\linewidth}{!}{%
$\displaystyle
\begin{aligned}
& \mathcal{L}_{\mathrm{DPO}}(\pi_{\theta}; \pi_{\mathrm{ref}}) = - \mathbb{E}_{(x, y_c, y_r) \sim \mathcal{D}} \\
& \left[ \log \sigma \left( \beta_{kl} \log \frac{\pi_{\theta}(y_c \mid x)}{\pi_{\mathrm{ref}}(y_c \mid x)} - \beta_{kl} \log \frac{\pi_{\theta}(y_r \mid x)}{\pi_{\mathrm{ref}}(y_r \mid x)} \right) \right],
\end{aligned}
$%
}
\end{equation}
where $\beta_{kl}$ is a hyperparameter controlling the deviation from the reference policy, and $\sigma$ denotes the logistic function.
\section{Method}
\subsection{Preference data construction}
\subsubsection{Data collection}
\begin{figure}[t]
  \centering
  \includegraphics[width=\columnwidth]{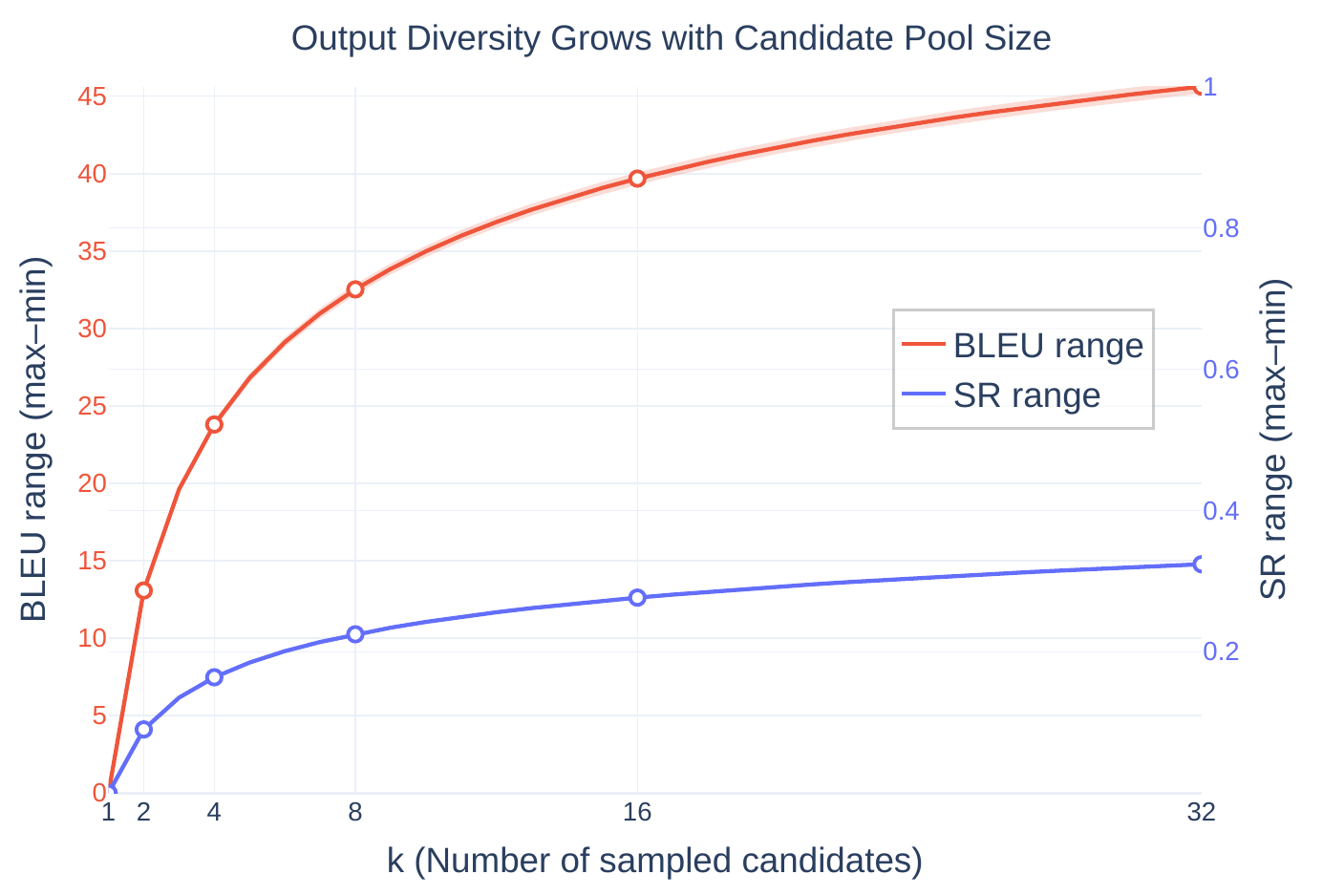}
\caption{Diversity as a function of candidate pool size $k$, measured by the range (max--min) of BLEU scores and silence ratios across $k$ candidates generated for the same query. The ranges grow with $k$ and plateau around $k=32$.}
  \label{fig:candidate_diversity}
\end{figure}

In order to construct an offline preference dataset for our model, we design a data collection pipeline consisting of three stages:

\begin{itemize}
    \item \textbf{Source data selection:} A mix of short- and long-form speech data was used to cover temporal diversity. For short-form instances ranging from 0 to 10 seconds, 10,000 utterances were randomly sampled from the CVSS-C~\cite{cvss_jia_2022} training set. For long-form instances spanning 10 to 60 seconds, 6,000 snippets were constructed from the French-to-English mTEDx~\cite{mtedx_salesky_2021} dataset by concatenating consecutive segments from the same TED talks.

    \item \textbf{Candidate generation:} To ensure linguistic diversity in the translation outputs, multiple candidate translations were sampled for each source French utterance using a decoding temperature of $1.0$. The number of samples was varied, and diversity was measured using the ranges (max--min) of BLEU scores and silence ratios across sampled candidates. As shown in Figure~\ref{fig:candidate_diversity}, both ranges increased with the number of samples and plateaued around $k=32$. Based on this observation, 32 candidates were generated per utterance in our experiments.

    \item \textbf{Quality and fluency measurement:} Before constructing the preference pairs, each candidate was evaluated across two primary dimensions: translation quality and acoustic fluency. To assess translation quality, Automatic Speech Recognition (ASR) was performed on the translated audio using the $\texttt{Whisper-medium}$~\cite{radford2022robustspeechrecognitionlargescale} model, and BLEU scores were subsequently computed between the recognized text and the ground-truth translation. For acoustic fluency, the silence ratio was quantified as the total duration of silence divided by the total speech duration from onset to offset. This ratio was computed using Silero VAD~\cite{silero_vad_2024} with the following default hyperparameters: $\texttt{VAD\_THRESHOLD}=0.5$, $\texttt{MIN\_SPEECH\_DURATION\_MS}=250$, and $\texttt{MIN\_SILENCE\_DURATION\_MS}=100$.
\end{itemize}

\subsubsection{Silver-Medal Preference}
\begin{figure}[t]
  \centering
  \includegraphics[width=\columnwidth]{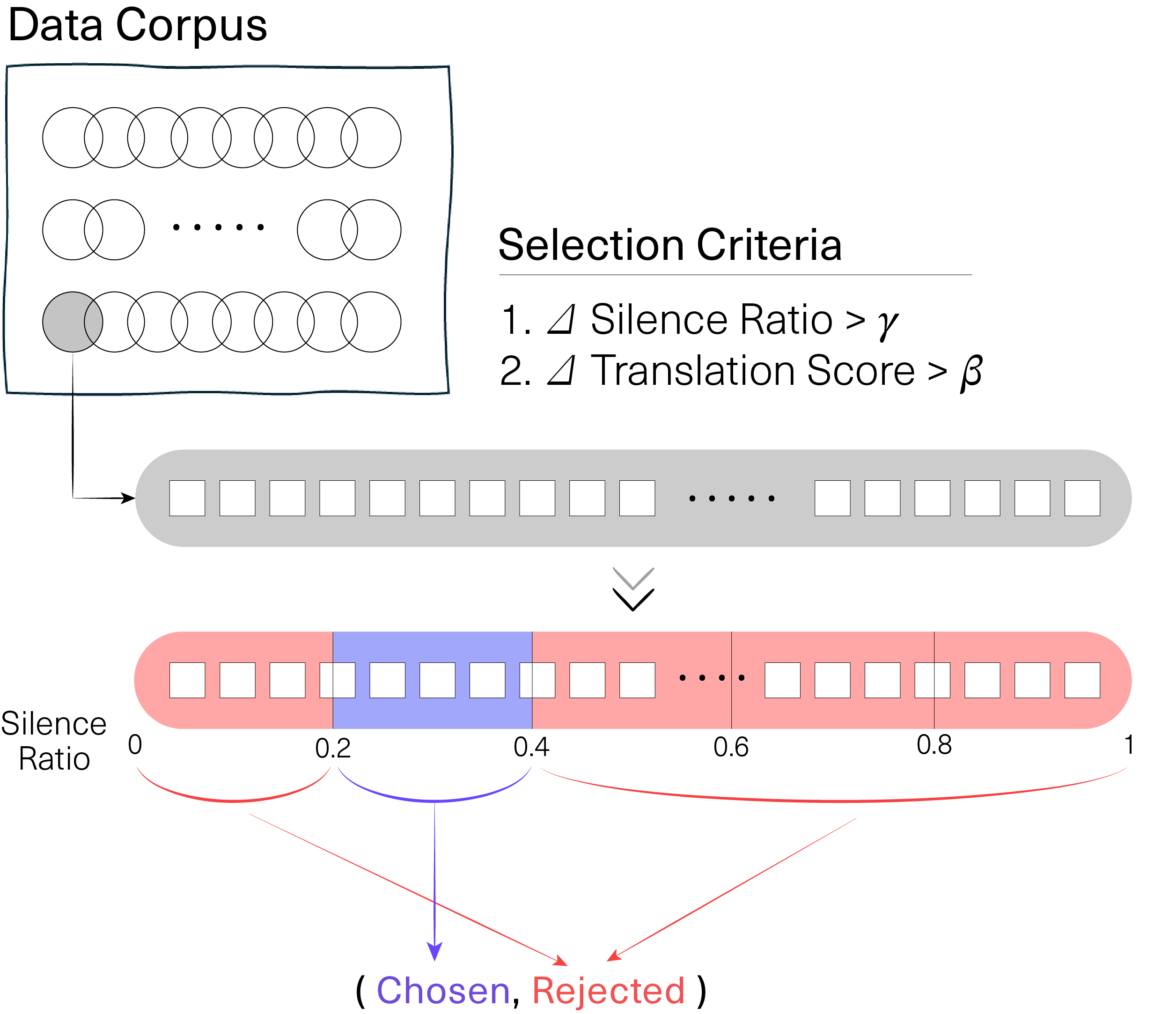}
  \caption{Illustration of the preference dataset construction process. The chosen samples are selected exclusively from the second-highest tier, rather than the top tier. The dataset is constructed so that the chosen and rejected samples differ by at least a predefined margin in both translation score and silence ratio.}
  \label{fig:preference_data}
\end{figure}

Here, as illustrated in Figure~\ref{fig:preference_data}, we describe how preference pairs are constructed.

\begin{itemize}

\item \textbf{Objective conflict in standard preference optimization:} The core mechanism of Direct Preference Optimization (DPO) relies on increasing the relative log-likelihood of chosen samples over rejected ones. Intuitively, to maximize fluency, one might strictly define candidates with the lowest silence ratios and high translation quality scores as the chosen set. However, blindly optimizing for a single axis induces a severe objective misalignment. Driven to eliminate silence, the model aggressively deviates from the ground-truth text to force unnatural acoustic continuity. This over-optimization neglects semantic fidelity, inevitably leading to a catastrophic degradation in translation quality as the model trades accurate meaning for unbroken speech.

\item \textbf{Silver-medal selection strategy:} To systematically prevent this collapse and ensure safe exploration, we stratify the 32 candidates into five equal quintiles based on their silence ratios. Counter-intuitively, we deliberately bypass the top-ranked tier and designate the second quintile (the top 20--40\%) as the explicitly chosen set. The core intention behind this ``silver-medal'' approach is to prevent the model from aggressively minimizing silence ratio, which inevitably leads to fluency overfitting and semantic degradation. By assigning both the overly aggressive first quintile and the third through fifth quintiles to the rejected pool, we bound the optimization space. This enables the model to stably explore the room for minimizing silence ratio within the 20--40\% range.

\item \textbf{Large-margin enforcement for optimization:} As a supplementary mechanism to guarantee distinct directional learning signals within this bounded space, we introduce strict margin requirements. Following~\cite{deng2026moreimprovingllmalignment}, which demonstrates that preference optimization benefits significantly more from candidate pairs with large margins than those with marginal differences, we formulate the optimization to require explicit performance gaps. Specifically, a chosen candidate must outperform its rejected counterpart by a translation quality margin $\beta$ and a silence-ratio margin $\gamma$. In our configuration, we define $\beta$ as a BLEU score difference of $5$, and $\gamma$ as $15\%$ of the group's normalized silence ratio. This large margin ensures that the preference gradients are driven only by unambiguous, high-quality improvements.
\end{itemize}

\subsection{Text-guided preference optimization for silence control}
\begin{itemize}

\item \textbf{Acoustically grounded stream isolation:} 
In preference optimization, a preference pair $(x, y_c, y_r)$ consists of a preceding source context $x$, a chosen translation $y_c$, and a rejected translation $y_r$. For $y \in \{y_c, y_r\}$, the translation trajectory $y$ intertwines the model's generated target text stream $T^y$, the target audio stream $A^y$, and the concurrently arriving source audio stream $S^y$. 

We define the joint probability of the target streams given the source context as:
$$ \pi_{\theta}(y \mid x) = \pi_{\theta}(A^y \mid x, S^y) \cdot \pi_{\theta}(T^y \mid x, A^y, S^y). $$

Early experiments and prior works \cite{wu2025aligningspokendialoguemodels} show that directly optimizing the raw acoustic token probabilities $\pi_{\theta}(A^y \mid \cdot)$ leads to highly unstable training. Ideally, one would marginalize over the high-cardinality discrete acoustic space $A^y$, but this is computationally intractable. Instead, we isolate the DPO objective exclusively to the text stream. Because our text policy $\pi_{\theta}(T^y \mid x, A^y, S^y)$ is auto-regressively conditioned on both the concurrent source stream $S^y$ and the generated target audio $A^y$, it inherently encapsulates the acoustic flow, latency delays, and silence ratios. By formulating our alignment target over this acoustically grounded text policy, $\pi_{\theta}^{T}(y \mid x) := \pi_{\theta}(T^y \mid x, A^y, S^y)$, we establish a stable and tractable optimization objective without losing the crucial streaming acoustic context.

\item \textbf{Length-normalized DPO for fluency:} Using the text logits computed from these sampled trajectories, we apply Length-Normalized Direct Preference Optimization (DPO-LN) to the chosen ($y_c$) and rejected ($y_r$) candidates selected by our preference selection strategy. Normalizing the log-probabilities by the generated text sequence length ($|T^{y_c}|$ and $|T^{y_r}|$) is crucial to prevent the model from unfairly penalizing longer, semantically accurate translations, thereby avoiding extreme fluency traps. The final objective is defined as:

\begin{equation}
\label{eq:dpo_ln_text}
\resizebox{0.9\linewidth}{!}{%
$\displaystyle
\begin{aligned}
& \mathcal{L}_{\mathrm{DPO-LN}}^{T}(\pi_{\theta}; \pi_{\mathrm{ref}}) = - \mathbb{E}_{(x, y_c, y_r) \sim \mathcal{D}} \\
& \left[ \log \sigma \left( \frac{\beta}{|T^{y_c}|} \log \frac{\pi_{\theta}^{T}(y_c \mid x)}{\pi_{\mathrm{ref}}^{T}(y_c \mid x)} - \frac{\beta}{|T^{y_r}|} \log \frac{\pi_{\theta}^{T}(y_r \mid x)}{\pi_{\mathrm{ref}}^{T}(y_r \mid x)} \right) \right],
\end{aligned}
$%
}
\end{equation}
\end{itemize}

\begin{table*}[t]
  \caption{Performance comparison of streaming speech translation models on Fr$\rightarrow$En benchmarks. 
NaturalFlow achieves competitive translation quality while maintaining low latency across both short-form (CVSS-C, VoxPopuli) and long-form (Audio-NTREX, mTEDx) datasets. Human speech denotes the original French speech from human speakers, with its silence ratio reported as a reference.}
  \label{tab:main_results_short}
  \centering

  \begin{subtable}[t]{\textwidth}
    
    \centering
    \resizebox{\textwidth}{!}{
      \begin{tabular}{ l cccccc cccccc }
        \toprule
        \multirow{3}{*}{Model} & \multicolumn{12}{c}{SHORT-FORM (FR-EN)} \\
        \cmidrule(lr){2-13}
        & \multicolumn{6}{c}{CVSS-C} & \multicolumn{6}{c}{VoxPopuli} \\
        \cmidrule(lr){2-7} \cmidrule(lr){8-13}
        & \makecell{SR ($\downarrow$)\footnotemark[1]} & LAAL ($\downarrow$) & \makecell{Start \\ Offset ($\downarrow$)} & \makecell{End \\ Offset ($\downarrow$)} & \makecell{ASR \\ BLEU ($\uparrow$)} & \makecell{ASR \\ COMET ($\uparrow$)}
        & \makecell{SR ($\downarrow$)} & LAAL ($\downarrow$) & \makecell{Start \\ Offset ($\downarrow$)} & \makecell{End \\ Offset ($\downarrow$)} & \makecell{ASR \\ BLEU ($\uparrow$)} & \makecell{ASR \\ COMET ($\uparrow$)} \\
        \midrule
        Human Speech & 0.02 & - & - & - & - & - & 0.06 & - & - & - & - & - \\
        \midrule
        StreamSpeech & 0.25 (0.49) & \textbf{1.96} & \textbf{1.97} & \textbf{1.77} & 22.15 & 0.61 & 0.36 & \textbf{2.26} & \textbf{1.31} & -1.26 & 7.06 & 0.34 \\
        Seamless     & 0.14 (0.42) & 3.24 & 3.40 & 2.85 & \textbf{35.74} & \textbf{0.83} & 0.27 & 3.65 & 2.67 & 5.05 & \textbf{19.74} & \textbf{0.73} \\
        Hibiki       & 0.08 (0.24) & 3.65 & 3.84 & 3.17 & 30.25 & 0.77 & 0.12 & 3.54 & 3.23 & 3.15 & 19.18 & 0.73 \\
        \midrule
        NaturalFlow  & \textbf{0.08 (0.11)} & 3.46 & 3.33 & 2.96 & 25.30 & 0.70 & \textbf{0.10} & 3.36 & 2.70 & \textbf{3.03} & 17.40 & 0.66 \\
        \bottomrule
      \end{tabular}
    }
  \end{subtable}

  \vspace{0.6em}

  \begin{subtable}[t]{\textwidth}
    \label{tab:main_results_long}
    \centering
    \resizebox{\textwidth}{!}{
      \begin{tabular}{ l cccccc cccccc }
        \toprule
        \multirow{3}{*}{Model} & \multicolumn{12}{c}{LONG-FORM (FR-EN)} \\
        \cmidrule(lr){2-13}
        & \multicolumn{6}{c}{Audio-NTREX} & \multicolumn{6}{c}{mTEDx} \\
        \cmidrule(lr){2-7} \cmidrule(lr){8-13}
        & \makecell{SR ($\downarrow$)} & LAAL ($\downarrow$) & \makecell{Start \\ Offset ($\downarrow$)} & \makecell{End \\ Offset ($\downarrow$)} & \makecell{ASR \\ BLEU ($\uparrow$)} & \makecell{ASR \\ COMET ($\uparrow$)}
        & \makecell{SR ($\downarrow$)} & LAAL ($\downarrow$) & \makecell{Start \\ Offset ($\downarrow$)} & \makecell{End \\ Offset ($\downarrow$)} & \makecell{ASR \\ BLEU ($\uparrow$)} & \makecell{ASR \\ COMET ($\uparrow$)} \\
        \midrule
        Human Speech & 0.08 & - & - & - & - & - & 0.23 & - & - & - & - & - \\
        \midrule
        StreamSpeech & 0.60 & 4.51 & \textbf{1.22} & -18.24 & 0.76 & 0.20 & 0.62 & 5.47 & \textbf{1.92} & -14.74 & 1.13 & 0.21 \\
        Seamless     & 0.27 & 4.63 & 2.82 & 3.30 & \textbf{25.34} & \textbf{0.37} & 0.65 & 7.72 & 3.43 & 21.28 & 27.55 & 0.41 \\
        Hibiki       & 0.17 & 3.65 & 3.05 & 2.85 & 24.07 & 0.32 & 0.26 & 3.69 & 3.16 & 0.97 & 32.94 & \textbf{0.46} \\
        \midrule
        NaturalFlow  & \textbf{0.13} & \textbf{3.49} & 2.79 & \textbf{2.53} & 23.96 & 0.34 & \textbf{0.21} & \textbf{3.38} & 2.58 & \textbf{0.82} & \textbf{33.27} & \textbf{0.46} \\
        \bottomrule
      \end{tabular}
    }
  \end{subtable}
\end{table*}

\footnotetext[1]{Since SR values are near zero for most CVSS-C clips due to their short duration, we additionally report the mean SR computed over the top 25\% of clips ranked by Hibiki SR values (i.e., the highest-SR quartile), to better illustrate performance trends of each model.}

\section{Experiments}
\subsection{Experimental settings}

We employ the Hibiki-2B model as our base architecture, fine-tuning it via Low-Rank Adaptation (LoRA)~\cite{hu2021loralowrankadaptationlarge} with a rank of $r=128$. We set text padding weight to 0.5 and the duration to 102.4. For preference alignment, we apply Direct Preference Optimization with length normalization utilizing a KL penalty coefficient of $\beta_{kl}=0.1$. The model is trained for 400 steps with an effective batch size of 32. We set the peak learning rate to $2\cdot10^{-6}$, accompanied by a 5\% one-cycle warmup schedule. All experiments were conducted on 4 NVIDIA L40S and 2 NVIDIA RTX PRO 6000 Blackwell GPUs.

\subsection{Baselines}
\subsubsection{Models}
We compare NaturalFlow against representative simultaneous S2ST baselines:
SeamlessStreaming~\cite{seamless_communication_2023} (Seamless), StreamSpeech~\cite{streamspeech_zhang_2024}, and Hibiki~\cite{hibiki_labiausse_2025}.
SeamlessStreaming is a multilingual streaming S2ST system that incrementally emits target speech as the source audio arrives.
StreamSpeech performs simultaneous S2ST with multi-task learning, jointly optimizing translation and online emission behavior for low-latency speech generation.
Hibiki is a high-fidelity direct S2ST model that discretizes audio with a neural codec and jointly generates target audio tokens with a temporally aligned text stream, operating in a fully continuous manner without an external control policy.

\subsubsection{Inference settings}
For SeamlessStreaming~\cite{seamless_communication_2023} and StreamSpeech~\cite{streamspeech_zhang_2024}, we run inference with SimulEval \cite{simuleval2020}, which provides (i) a delay-free waveform that concatenates emitted target speech segments and (ii) a delay log that records the emission timestamp of each segment relative to the source audio.

However, since this waveform omits the inter-chunk silences that would be present during real-time streaming, directly applying VAD and ASR would yield timestamps misaligned with the source audio timeline $t=0$. To correct for this, we reconstruct the original emission timeline by re-inserting silence intervals into the waveform according to the delay log, prior to running Silero VAD~\cite{silero_vad_2024} for segment boundaries and WhisperX~\cite{whisperx_bain_2023} for word-level timestamps and transcriptions. This reconstruction ensures that inter-chunk silences are preserved and that all systems are evaluated under the same pipeline, enabling a fair comparison of latency and fluency metrics.
\section{Evaluation}
\subsection{Benchmarks}
\subsubsection{Short-form data}
\begin{itemize}
    \item \textbf{CVSS-C}: We use the Fr-En \textit{test} split of CVSS-C~\cite{cvss_jia_2022}, a widely used S2ST benchmark derived from Common Voice~\cite{common_voice_ardila_2020} recordings with paired translation text from CoVoST~2~\cite{covost_wang_2021}. This benchmark contains real-speaker French source audio with an average duration of $5.6$\,s.

    \item \textbf{VoxPopuli S2S Interpretation}: We use the Fr-En Speech-to-Speech Interpretation subset of VoxPopuli~\cite{voxpopuli_wang_2021}, from which we randomly sample 1,000 examples for our test set. This benchmark is derived from European Parliament speech and consists of real French source audio with an average duration of $11.4$\,s.
\end{itemize}

\subsubsection{Long-form data}
\begin{itemize}
    \item \textbf{Audio-NTREX-4L}: We utilize the Fr-En \textit{test} split of Audio-NTREX-4L, a multilingual long-form ST dataset proposed in Hibiki-zero~\cite{hibiki_zero_labiausse_2026}. This benchmark is constructed by synthesizing text translations from the NTREX corpus~\cite{federmann-etal-2022-ntrex} using high-quality industry TTS engines. The samples used in our evaluation have an average duration of $42.1$\,s.
    
    \item \textbf{mTEDx}: We use the Fr-En portions of mTEDx~\cite{mtedx_salesky_2021} \textit{test} and \textit{validation} splits, which are originally provided at the segment level, to construct our own \textit{test} split. To construct long-form evaluation instances, we concatenate consecutive segments that are semantically contiguous, forming audio sequences with total duration between 20 and 60 seconds. The resulting benchmark has an average duration of $35.8$\,s and comprises real spoken lectures with sustained discourse, enabling evaluation of models under realistic long-context speaking patterns.
\end{itemize}

\subsection{Metrics}
\subsubsection{Translation quality}
\begin{itemize}
    \item \textbf{ASR-BLEU.}
    For each system output, we transcribe the generated speech using \texttt{Whisper-medium} \cite{radford2022robustspeechrecognitionlargescale}.
    We then compute BLEU between the ASR transcript and the ground-truth (GT) translation text provided by each benchmark.
    We use \texttt{SacreBLEU} \cite{sacre_bleu_post_2018} for standardized BLEU computation.

    \item \textbf{ASR-COMET.}
    Using the same Whisper-medium transcripts, we evaluate reference-based translation quality with XCOMET-XL \cite{xcomet_guerreiro_2023},
    where the benchmark GT translation is used as the reference.
\end{itemize}

\subsubsection{Latency}
We report two offset-based latency metrics (Start Offset and End Offset), following the simultaneous speech-to-speech translation evaluation protocol of IWSLT~\cite{iwslt_ahmad-etal-2024-findings}, and one lagging-based metric~\cite{laal_papi_2022}.

Let $X=[x_1,\ldots,x_{|X|}]$ be the sequence of \emph{source speech segments} and
$Y=[y_1,\ldots,y_{|Y|}]$ be the sequence of \emph{output speech segments}, both extracted by Silero VAD~\cite{silero_vad_2024}.
For each segment $z$, let $(t^{s}(z), t^{e}(z))$ denote its start and end timestamps.
Throughout, timestamps are measured on the same time axis, where the source audio starts at time $0$.

\textbf{Start Offset.}
Start offset measures how long the system stays silent before producing the first audible output speech segment, defined as the onset time of the first output segment on the global timeline, i.e., $\mathrm{StartOffset} = t^{s}(y_1)$.
Since all timestamps are measured on the same axis where the source audio starts at time $0$, this quantity directly represents the elapsed time from the beginning of the source audio until the first output speech begins.

\textbf{End Offset.}
Similarly, end offset measures how long the system continues speaking after the source speech has ended, defined as $t^{e}(y_{|Y|}) - t^{e}(x_{|X|})$.

\textbf{Length-Adaptive Average Lagging (LAAL).}
Following \cite{laal_papi_2022}, LAAL estimates the average source--target delay without explicit word-level alignment.
Let $D_{\mathrm{src}}$ be the source-speech duration.
We obtain word-level timestamps for the output transcripts using WhisperX \cite{whisperx_bain_2023}, yielding a sequence of \emph{generated word segments}
$W=[w_1,\ldots,w_{|W|}]$ on the same time axis, where each word segment $w_i$ has timestamps
$\bigl(t^{s}(w_i),\,t^{e}(w_i)\bigr)$.
We use the word \emph{emission time} $d_i \triangleq t^{s}(w_i)$.

Let $n_{\mathrm{gen}}=|W|$ be the number of generated words and let $n_{\mathrm{ref}}$ be the number of words in the reference translation.
LAAL defines the oracle emission schedule using the \emph{maximum} length to avoid rewarding under-/over-generation \cite{laal_papi_2022}:
$d^{*}_i = (i-1)\cdot \frac{D_{\mathrm{src}}}{\max\left(n_{\mathrm{gen}},\,n_{\mathrm{ref}}\right)}.$
Let $\tau \triangleq \min\{\, i \mid d_i \ge D_{\mathrm{src}} \,\}$ be the first word index emitted at or after the end of the source speech.
Then LAAL is computed as $\mathrm{LAAL} = \frac{1}{\tau}\sum_{i=1}^{\tau}\left(d_i - d^{*}_i\right).$
Intuitively, LAAL averages how much the actual emissions $d_i$ lag behind an ``oracle'' that spreads word emissions uniformly over $D_{\mathrm{src}}$
using $\max(n_{\mathrm{gen}},n_{\mathrm{ref}})$, making the metric robust to both under- and over-generation.

\subsubsection{Fluency}
As a measure of fluency, we quantify the degree of silence in generated speech using Silence Ratio (SR).

\textbf{Silence Ratio (SR).}
We propose Silence Ratio as a fluency metric that quantifies the fraction of time the system is \emph{silent} within the span of its emitted speech.
Using the output speech segments $Y$ from Silero VAD, we define the output speech span
\begin{equation}
\label{eq:sr_span}
D_{\mathrm{span}} \;=\; t^{e}(y_{|Y|}) - t^{s}(y_1),
\end{equation}
and the total voiced duration
\begin{equation}
\label{eq:sr_voiced}
D_{\mathrm{voice}} \;=\; \sum_{j=1}^{|Y|}\left(t^{e}(y_j) - t^{s}(y_j)\right).
\end{equation}
The internal silence duration is $D_{\mathrm{sil}} = D_{\mathrm{span}} - D_{\mathrm{voice}}$, and we compute
\begin{equation}
\label{eq:sr}
\mathrm{SR} = \frac{D_{\mathrm{sil}}}{D_{\mathrm{span}}} = 1 - \frac{D_{\mathrm{voice}}}{D_{\mathrm{span}}}.
\end{equation}
With our Silero VAD configuration, we adopt a 0.10\,s minimum silence cutoff (\texttt{MIN\_SILENCE\_DURATION\_MS}=100). This choice follows established practice in pause measurement, where the minimum cut-off is commonly set around 0.10--0.28\,s \cite{Mead2005}. In such setups, pauses shorter than the cut-off are typically excluded from pause-duration calculations~\cite{Mead2005}. Accordingly, in our SR computation, gaps shorter than 0.10\,s do not terminate a speech segment and are therefore not counted as silence, making SR less sensitive to brief natural gaps (e.g., short inter-word pauses).

\subsection{Human evaluation}
To assess whether our model produces more natural translations than the baseline, we conducted a human evaluation using Amazon Mechanical Turk (MTurk)~\cite{mturk_snow_2008}. Since over half of the samples in the CVSS-C test set naturally exhibit a silence ratio of $0$ for both systems, evaluating the entire set would dilute the comparison of fluency improvements. Therefore, we randomly sampled instances from the top 25\% of the baseline SR distribution. From this subset, we provided annotators with paired translation audio and asked them to select their preferred translation based on overall naturalness and fluency. We collected judgments from 30 independent raters, yielding a total of approximately 150 human ratings.

\section{Results}
In our experiments, we aim to answer the following three research questions. First, we investigate whether our method can effectively reduce the silence ratio without incurring a trade-off in translation quality. Second, we analyze the stability of our preference design to ensure it does not collapse into a single objective, such as exclusively minimizing silence or solely maximizing translation scores. We compare its behavior against standard preference settings, which push the model toward the extremes of each objective—either aggressively minimizing silence or maximizing translation scores. Finally, to ensure our objective improvements align with human perception, we conduct human evaluations focusing on the naturalness and fluency of the generated speech.

\begin{figure}[t]
  \centering
  \includegraphics[width=\columnwidth]{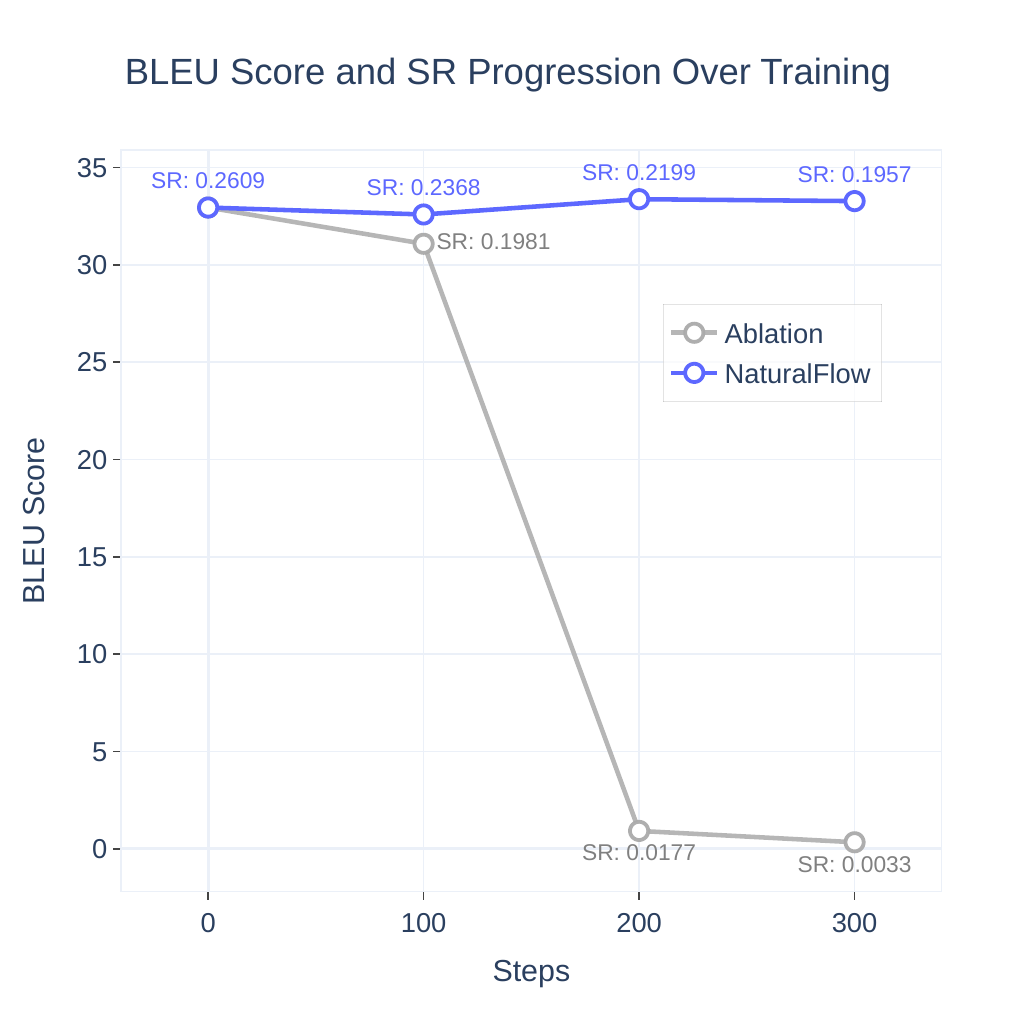}
  \caption{\textbf{Ablation 1.} Progression of BLEU score and silence ratio over training steps on the mTEDx test set. The plot demonstrates that our model robustly maintains its translation performance even as the silence ratio decreases, whereas the ablation model suffers a severe drop in BLEU score.}
  \label{fig:SR_BLEU}
\end{figure}

\begin{figure}[h]
  \centering
  \includegraphics[width=\columnwidth]{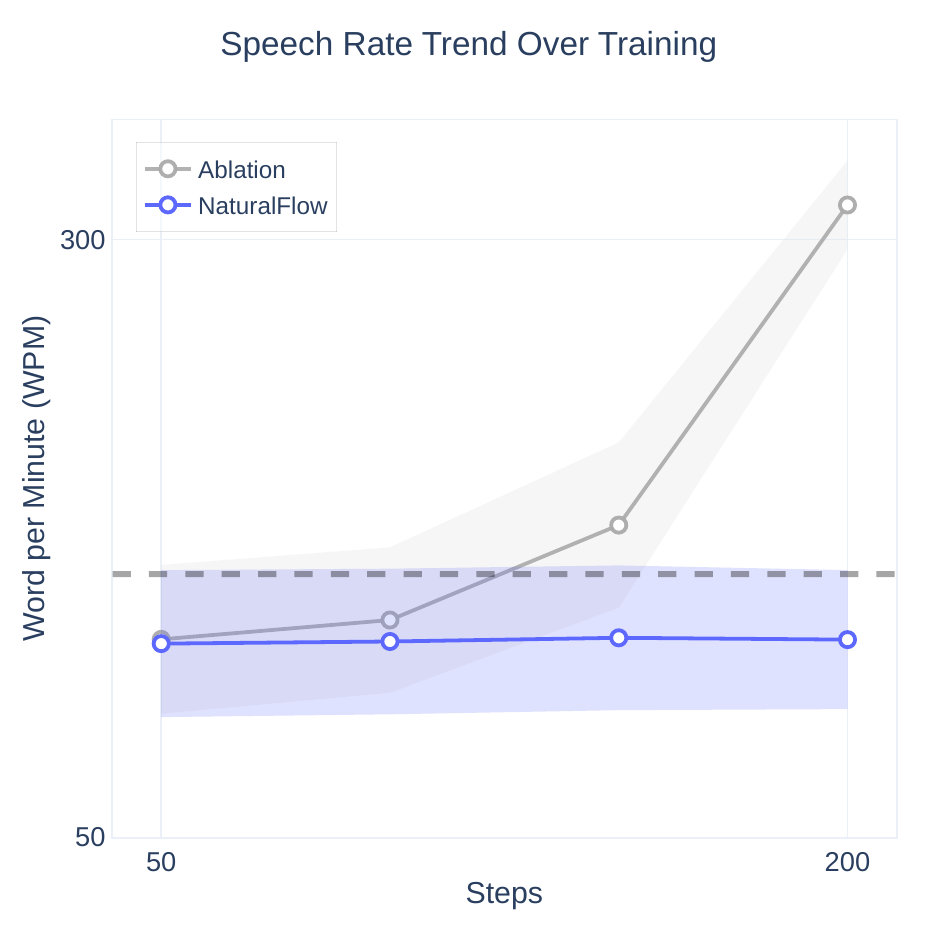}
  \caption{\textbf{Ablation 2.} Comparison of speech speed between our model and a model trained with preference data without the low-SR group. Removing the low-SR group from preference data significantly reduces the silence ratio, leading to excessively fast speech. The dashed gray line indicates the typical average human speaking rate (160 words per minute~\cite{rayner2008language}).}
  \label{fig:WPM}
\end{figure}

\subsection{Silence reduction with preservation of other metrics}
Table~\ref{tab:main_results_short} summarizes the main trade-off between output continuity, translation quality, and latency. Across short-form and long-form benchmarks, NaturalFlow reduces SR relative to the baseline while keeping translation quality in a comparable range, without substantially degrading latency-related metrics.

On short-form benchmarks, NaturalFlow matches the lowest average SR on CVSS-C and attains the lowest SR on VoxPopuli. On CVSS-C, it achieves a similar average SR (0.08) to the baseline, while reducing SR on the top 25\% high-SR subset ranked by Hibiki's SR. On VoxPopuli, it reduces SR from $0.12$ (Hibiki) to $0.10$. Translation quality remains competitive: NaturalFlow is slightly below Hibiki in ASR-BLEU/COMET on short-form, but the gap is moderate. Latency metrics (LAAL, start/end offsets) remain in the same range as baselines, indicating that SR reduction is not achieved by a latency increase.

\begin{figure}[t]
  \centering
  \includegraphics[width=\columnwidth]{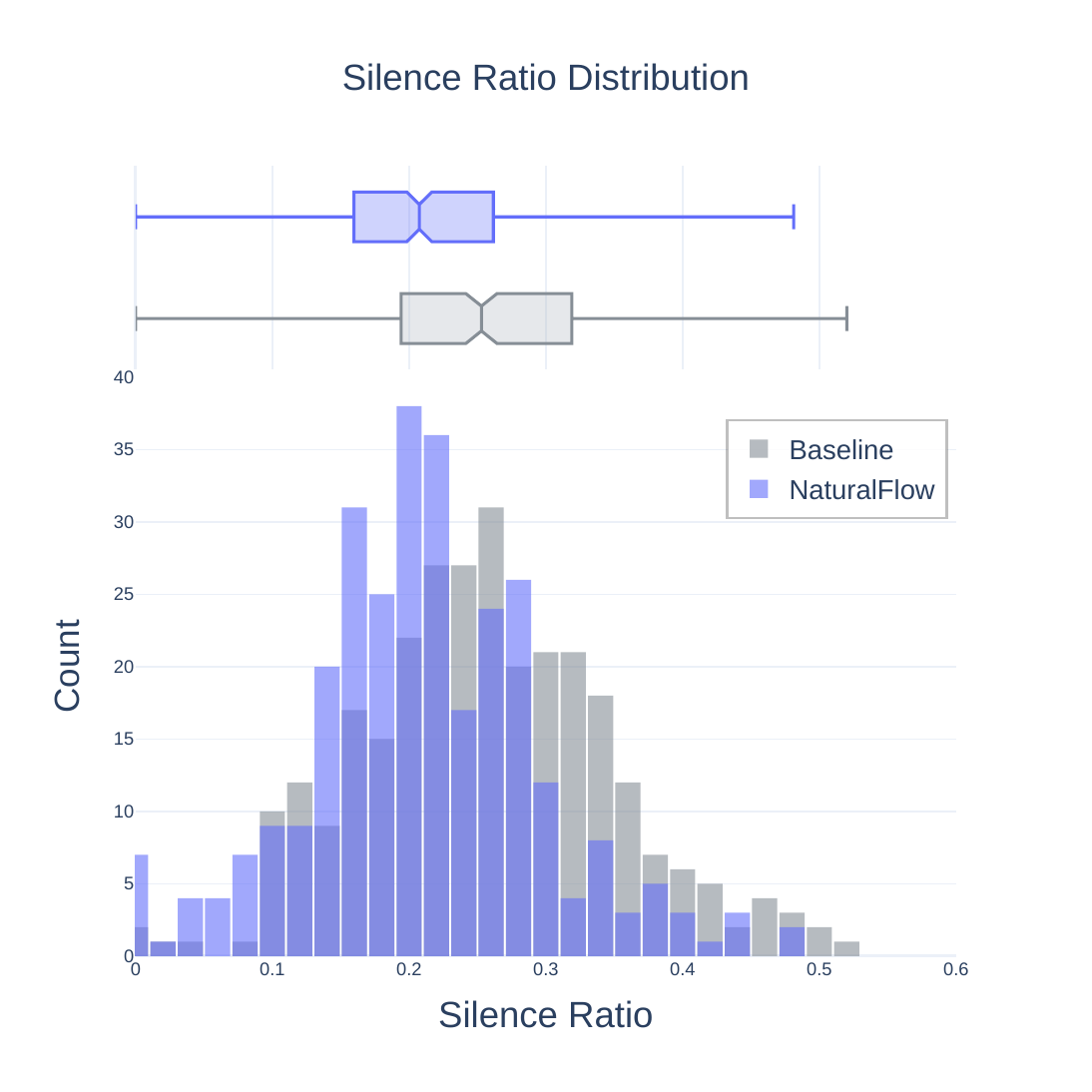}
  \caption{Silence-ratio distribution shift on the mTEDx test set.}
  \label{fig:SR_dist}
\end{figure}

On long-form benchmarks, our model also achieves the lowest SR on Audio-NTREX ($0.13$) and mTEDx ($0.21$) (Figure~\ref{fig:SR_dist}). At the same time, it maintains translation quality, reaching even slightly better ASR-BLEU on mTEDx and comparable COMET to the best model. Our model also matches or improves latency metrics in this setting (best LAAL on both datasets and competitive boundary offsets), showing no evidence of a latency-quality regression. Overall, these results support our claim that SR can be reduced without compromising translation quality, encouraging the model to explore the trade-off between silence ratio and translation quality.

\subsection{Ablation study: stability of Silver-Medal preference design}
To validate the stability and necessity of our Silver-Medal preference design, we conduct an ablation study on the preference dataset construction. We specifically investigate the impact of removing our group-constrained mechanisms, which are designed to explicitly prevent the model from over-optimizing the silence ratio at the expense of translation quality.

First, we compare our approach against a \textbf{Standard Setting (Ablation 1)}. In this setup, we construct the chosen set strictly from the top 20\% of candidates with the lowest silence ratio, while randomly sampling the rejected candidates from the remaining pool. To ensure a strong directional learning signal, we enforced a strict margin: a silence ratio gap of at least 0.2 between the chosen and rejected pairs, and a requirement that the chosen candidates must exceed the baseline average ASR-BLEU score.

Second, we test \textbf{Removing the Low-SR Band (Ablation 2)}. The core intuition of our Silver-Medal strategy is to actively penalize overly aggressive, low-SR samples by placing the extreme low-SR group into the rejected group. In this ablation, we remove this low-SR penalty to explicitly test whether this mechanism is crucial for preventing the model from over-optimizing the silence ratio.

As described in Table~\ref{tab:evaluation_results} and illustrated by Figures~\ref{fig:SR_BLEU} and~\ref{fig:WPM}, both ablated settings suffer from severe optimization collapse. While the models rapidly decrease the silence ratio as training progresses, this reduction comes at the expense of substantial degradation in translation quality. Specifically, observing the behavioral patterns of these models reveals that this one-sided optimization leads to an excessively fast speaking rate; as shown in Figure \ref{fig:WPM}, the model attempts to eliminate pauses almost entirely, generating non-stop, extremely fast speech that becomes unintelligible. Consequently, the model is left with no opportunity to balance acoustic fluency and semantic accuracy.

These findings confirm the critical role of our Silver-Medal strategy. Without explicitly bounding the optimization space—particularly by using the extreme low-SR band as a negative signal—the model easily falls into single-objective optimization. By preventing this acceleration, our preference design successfully stabilizes training, enabling a reduction in unnatural pauses while preserving robust translation quality.

\subsection{Human evaluation on naturalness}
While automated metrics indicate a successful reduction in silence ratio, we conducted subjective human evaluations to verify whether these objective gains translate to a genuinely improved listening experience. Evaluators were presented with paired audio samples and asked to choose the interpretation they perceived as more natural and acoustically fluent. 

The evaluation results, summarized in Table \ref{tab:human_eval_naturalness}, demonstrate the effectiveness of our fluency-aware optimization. First, our NaturalFlow model received a higher preference rate than the baseline system (55\% vs. 34\%). This confirms our primary hypothesis: explicitly targeting the reduction of unnatural pauses enhances the perceived naturalness of the translated speech. 

Furthermore, we evaluate the necessity of our Silver-Medal strategy through an ablation study. Human evaluators showed a strong preference for the NaturalFlow framework (68\% vs. 24\%) over the ablated model. Crucially, without our preference dataset construction design, optimization collapses toward a single objective of minimizing the silence ratio, leading to an unnatural, rushed delivery. These results indicate that our constrained selection is essential for bypassing this fluency trap, successfully ensuring acoustic continuity with a comfortable listening experience.

\begin{table}[t]
  \caption{Ablation study on our preference dataset construction method on CVSS-C and mTEDx datasets.}
  \label{tab:evaluation_results}
  \centering
  \resizebox{\columnwidth}{!}{%
  \begin{tabular}{l ccc ccc}
    \toprule
    & \multicolumn{3}{c}{\textbf{CVSS-C}} & \multicolumn{3}{c}{\textbf{mTEDx}} \\
    \cmidrule(lr){2-4} \cmidrule(lr){5-7}
    \textbf{Models} & SR & ASR-BLEU & ASR-COMET & SR & ASR-BLEU & ASR-COMET \\
    \midrule
    Ours & 0.11 & 25.30 & 0.70 & 0.21 & 33.27 & 0.46 \\
    Ablation 1 & 0.01 & 1.50 & 0.26 & 0.00 & 1.41 & 0.18 \\
    Ablation 2 & 0.00 & 1.60 & 0.25 & 0.00 & 0.92 & 0.17 \\
    \bottomrule
  \end{tabular}%
  }
\end{table}

\begin{table}[ht!]
  \caption{Human evaluation results on interpretation naturalness. Values represent the percentage of responses in which evaluators preferred NaturalFlow (Pref.), perceived no difference (Tie), or preferred the compared model (Unpref.). All comparisons are statistically significant by a binomial test $(p < 0.05).$}
  \label{tab:human_eval_naturalness}
  \centering
  \small
  \begin{tabular}{l ccc}
    \toprule
    \textbf{Compared Model} & \textbf{Pref.} & \textbf{Tie} & \textbf{Unpref.} \\
    \midrule
    Baseline & 55 \% & 11 \% & 34 \% \\
    w/o Silver-Medal Preference & 68 \% & 8 \% & 24 \% \\
    \bottomrule
  \end{tabular}
\end{table}

\section{Conclusion}
In this work, we present a fluency-aware optimization framework for simultaneous speech-to-speech translation (S2ST) that reduces unnatural pauses while maintaining translation fidelity. By integrating the silence ratio as an optimization objective with our Silver-Medal Preference design, we balance the continuity of speech flow with translation quality, preventing the model from collapsing toward only a single objective. Experimental results, supported by human assessment, demonstrate that our method lowers the silence ratio while preserving standard translation and latency metrics. We believe this work lays crucial groundwork for advancing delivery naturalness in S2ST, serving as a valuable foundation for extending this framework across a broader range of language pairs.

\section{Acknowledgements}
This work was supported by Institute of Information \& Communications Technology Planning \& Evaluation (IITP) grants funded by the Korean government (MSIT) [NO.RS-2021-II211343, Artificial Intelligence Graduate School Program (Seoul National University); No.2022-0-00959, RS-2022-II220959], by National Research Foundation of Korea (NRF) grant [No.2022R1A3B1077720, 2022R1A5A7083908], BK21 FOUR Program of the Education and Research Program for Future ICT Pioneers, Seoul National University in 2026. This was supported by Mobile eXperience(MX) Business, Samsung Electronics Co., Ltd. In addition, this work was supported by the 2026 Research Fund of the University of Seoul for Heeseung Kim.
\section{Generative AI Use Disclosure}
Generative AI tools were used solely for language editing and polishing. All scientific content, including the methodology, experiments, results, and conclusions, was developed and verified by the authors.
\bibliographystyle{IEEEtran}
\bibliography{mybib}

@article{Rennert2010,
  author = {Rennert, Sylvi},
  title = {The impact of fluency on the subjective assessment of interpreting quality},
  journal = {Interpreting},
  volume = {12},
  number = {1},
  pages = {1--24},
  year = {2010},
}

@article{rayner2008language,
title = {Language processing in reading and speech perception is fast and incremental: Implications for event-related potential research},
journal = {Biological Psychology},
volume = {80},
number = {1},
pages = {4-9},
year = {2009},
note = {Before the N400: Early Latency Language ERPs},
issn = {0301-0511},
doi = {https://doi.org/10.1016/j.biopsycho.2008.05.002},
url = {https://www.sciencedirect.com/science/article/pii/S0301051108001245},
author = {Keith Rayner and Charles Clifton}
}

@misc{hu2021loralowrankadaptationlarge,
      title={LoRA: Low-Rank Adaptation of Large Language Models}, 
      author={Edward J. Hu and Yelong Shen and Phillip Wallis and Zeyuan Allen-Zhu and Yuanzhi Li and Shean Wang and Lu Wang and Weizhu Chen},
      year={2021},
      eprint={2106.09685},
      archivePrefix={arXiv},
      primaryClass={cs.CL},
      url={https://arxiv.org/abs/2106.09685}, 
}

@misc{radford2022robustspeechrecognitionlargescale,
      title={Robust Speech Recognition via Large-Scale Weak Supervision}, 
      author={Alec Radford and Jong Wook Kim and Tao Xu and Greg Brockman and Christine McLeavey and Ilya Sutskever},
      year={2022},
      eprint={2212.04356},
      archivePrefix={arXiv},
      primaryClass={eess.AS},
      url={https://arxiv.org/abs/2212.04356}, 
}

@misc{ouyang2022traininglanguagemodelsfollow,
      title={Training language models to follow instructions with human feedback}, 
      author={Long Ouyang and Jeff Wu and Xu Jiang and Diogo Almeida and Carroll L. Wainwright and Pamela Mishkin and Chong Zhang and Sandhini Agarwal and Katarina Slama and Alex Ray and John Schulman and Jacob Hilton and Fraser Kelton and Luke Miller and Maddie Simens and Amanda Askell and Peter Welinder and Paul Christiano and Jan Leike and Ryan Lowe},
      year={2022},
      eprint={2203.02155},
      archivePrefix={arXiv},
      primaryClass={cs.CL},
      url={https://arxiv.org/abs/2203.02155}, 
}

@misc{deng2026moreimprovingllmalignment,
      title={Less is More: Improving LLM Alignment via Preference Data Selection}, 
      author={Xun Deng and Han Zhong and Rui Ai and Fuli Feng and Zheng Wang and Xiangnan He},
      year={2026},
      eprint={2502.14560},
      archivePrefix={arXiv},
      primaryClass={cs.LG},
      url={https://arxiv.org/abs/2502.14560}, 
}

@InProceedings{wu2025aligningspokendialoguemodels,
  title = 	 {Aligning Spoken Dialogue Models from User Interactions},
  author =       {Wu, Anne and Mazar\'{e}, Laurent and Zeghidour, Neil and D\'{e}fossez, Alexandre},
  booktitle = 	 {Proceedings of the 42nd International Conference on Machine Learning},
  pages = 	 {67476--67498},
  year = 	 {2025},
  editor = 	 {Singh, Aarti and Fazel, Maryam and Hsu, Daniel and Lacoste-Julien, Simon and Berkenkamp, Felix and Maharaj, Tegan and Wagstaff, Kiri and Zhu, Jerry},
  volume = 	 {267},
  series = 	 {Proceedings of Machine Learning Research},
  month = 	 {13--19 Jul},
  publisher =    {PMLR},
  url = 	 {https://proceedings.mlr.press/v267/wu25t.html}
}

@misc{lee2024rlaifvsrlhfscaling,
      title={RLAIF vs. RLHF: Scaling Reinforcement Learning from Human Feedback with AI Feedback}, 
      author={Harrison Lee and Samrat Phatale and Hassan Mansoor and Thomas Mesnard and Johan Ferret and Kellie Lu and Colton Bishop and Ethan Hall and Victor Carbune and Abhinav Rastogi and Sushant Prakash},
      year={2024},
      eprint={2309.00267},
      archivePrefix={arXiv},
      primaryClass={cs.CL},
      url={https://arxiv.org/abs/2309.00267}, 
}

@misc{schulman2017proximalpolicyoptimizationalgorithms,
      title={Proximal Policy Optimization Algorithms}, 
      author={John Schulman and Filip Wolski and Prafulla Dhariwal and Alec Radford and Oleg Klimov},
      year={2017},
      eprint={1707.06347},
      archivePrefix={arXiv},
      primaryClass={cs.LG},
      url={https://arxiv.org/abs/1707.06347}, 
}

@article{Mead2005,
  title={Methodological issues in the study of interpreters' fluency},
  author={Mead, Peter},
  year={2005},
  publisher={EUT-Edizioni Universit{\`a} di Trieste}
}

@article{Tissi2000,
  title={Silent pauses and disfluencies in simultaneous interpretation: A descriptive analysis},
  author={Tissi, Benedetta},
  journal={The Interpreters’ Newsletter},
  volume={10},
  number={4},
  pages={103--127},
  year={2000}
}

@article{Han2015,
  title={(Para) linguistic correlates of perceived fluency in English-to-Chinese simultaneous interpretation},
  author={Han, Chao},
  journal={International Journal of Comparative Literature \& Translation Studies},
  volume={3},
  number={4},
  pages={32},
  year={2015},
  publisher={Australian International Academic Centre PTY. Ltd (AIAC)}
}

@article{Gile1999, title={Testing the Effort Models’ tightrope hypothesis in simultaneous interpreting - A contribution}, volume={12}, url={https://tidsskrift.dk/her/article/view/25553}, DOI={10.7146/hjlcb.v12i23.25553}, abstractNote={In a sample of 10 professionals interpreting the same source speech in the simultaneous mode, errors and omissions (e/o’s) were found to affect different source-speech seg-ments, and a large proportion among them were only made by a small proportion of the subjects. In a repeat performance, there were some new e/o’s in the second version when the same interpreters had interpreted the same segments correctly in the first version. These findings strengthen the Effort Models’ “tightrope hypothesis” that many e/o’s are due not to the intrinsic difficulty of the corresponding source-speech segments, but to the interpreters working close to processing capacity saturation, which makes them vulnerable to even small variations in the available processing capacity for each interpreting component.}, number={23}, journal={HERMES - Journal of Language and Communication in Business}, author={Gile, Daniel}, year={1999}, month={Feb.}, pages={153–172} }

@inproceedings{Lennon2000,
  title={The lexical element in spoken second language fluency},
  author={P. Alan Lennon},
  year={2000},
  url={https://api.semanticscholar.org/CorpusID:151390810}
}

@inproceedings{jia2019,
  title     = {{Direct Speech-to-Speech Translation with a Sequence-to-Sequence Model}},
  author    = {Ye Jia and Ron J. Weiss and Fadi Biadsy and Wolfgang Macherey and Melvin Johnson and Zhifeng Chen and Yonghui Wu},
  year      = {2019},
  booktitle = {{Interspeech 2019}},
  pages     = {1123--1127},
  doi       = {10.21437/Interspeech.2019-1951},
  issn      = {2958-1796},
}

@inproceedings{lee2022,
    title = "Direct Speech-to-Speech Translation With Discrete Units",
    author = "Lee, Ann  and
      Chen, Peng-Jen  and
      Wang, Changhan  and
      Gu, Jiatao  and
      Popuri, Sravya  and
      Ma, Xutai  and
      Polyak, Adam  and
      Adi, Yossi  and
      He, Qing  and
      Tang, Yun  and
      Pino, Juan  and
      Hsu, Wei-Ning",
    editor = "Muresan, Smaranda  and
      Nakov, Preslav  and
      Villavicencio, Aline",
    booktitle = "Proceedings of the 60th Annual Meeting of the Association for Computational Linguistics (Volume 1: Long Papers)",
    month = may,
    year = "2022",
    address = "Dublin, Ireland",
    publisher = "Association for Computational Linguistics",
    url = "https://aclanthology.org/2022.acl-long.235/",
    doi = "10.18653/v1/2022.acl-long.235",
    pages = "3327--3339"
}

@inproceedings{weiss2017,
  title     = {{Sequence-to-Sequence Models Can Directly Translate Foreign Speech}},
  author    = {Ron J. Weiss and Jan Chorowski and Navdeep Jaitly and Yonghui Wu and Zhifeng Chen},
  year      = {2017},
  booktitle = {{Interspeech 2017}},
  pages     = {2625--2629},
  doi       = {10.21437/Interspeech.2017-503},
  issn      = {2958-1796},
}

@misc{défossez2024moshispeechtextfoundationmodel,
      title={Moshi: a speech-text foundation model for real-time dialogue}, 
      author={Alexandre Défossez and Laurent Mazaré and Manu Orsini and Amélie Royer and Patrick Pérez and Hervé Jégou and Edouard Grave and Neil Zeghidour},
      year={2024},
      eprint={2410.00037},
      archivePrefix={arXiv},
      primaryClass={eess.AS},
      url={https://arxiv.org/abs/2410.00037}, 
}

@article{cvss_jia_2022,
  author       = {Ye Jia and
                  Michelle Tadmor Ramanovich and
                  Quan Wang and
                  Heiga Zen},
  title        = {{CVSS} Corpus and Massively Multilingual Speech-to-Speech Translation},
  journal      = {CoRR},
  volume       = {abs/2201.03713},
  year         = {2022},
  url          = {https://arxiv.org/abs/2201.03713},
  eprinttype    = {arXiv},
  eprint       = {2201.03713},
  bibsource    = {dblp computer science bibliography, https://dblp.org}
}

@misc{mtedx_salesky_2021,
      title={The Multilingual TEDx Corpus for Speech Recognition and Translation}, 
      author={Elizabeth Salesky and Matthew Wiesner and Jacob Bremerman and Roldano Cattoni and Matteo Negri and Marco Turchi and Douglas W. Oard and Matt Post},
      year={2021},
      eprint={2102.01757},
      archivePrefix={arXiv},
      primaryClass={cs.CL},
      url={https://arxiv.org/abs/2102.01757}, 
}

@misc{voxpopuli_wang_2021,
      title={VoxPopuli: A Large-Scale Multilingual Speech Corpus for Representation Learning, Semi-Supervised Learning and Interpretation}, 
      author={Changhan Wang and Morgane Rivière and Ann Lee and Anne Wu and Chaitanya Talnikar and Daniel Haziza and Mary Williamson and Juan Pino and Emmanuel Dupoux},
      year={2021},
      eprint={2101.00390},
      archivePrefix={arXiv},
      primaryClass={cs.CL},
      url={https://arxiv.org/abs/2101.00390}, 
}

@inproceedings{simuleval2020,
    title = "{SIMULEVAL}: An Evaluation Toolkit for Simultaneous Translation",
    author = "Ma, Xutai  and
      Dousti, Mohammad Javad  and
      Wang, Changhan  and
      Gu, Jiatao  and
      Pino, Juan",
    editor = "Liu, Qun  and
      Schlangen, David",
    booktitle = "Proceedings of the 2020 Conference on Empirical Methods in Natural Language Processing: System Demonstrations",
    month = oct,
    year = "2020",
    address = "Online",
    publisher = "Association for Computational Linguistics",
    url = "https://aclanthology.org/2020.emnlp-demos.19/",
    doi = "10.18653/v1/2020.emnlp-demos.19",
    pages = "144--150"
}

@InProceedings{hibiki_labiausse_2025,
  title = 	 {High-Fidelity Simultaneous Speech-To-Speech Translation},
  author =       {Labiausse, Tom and Mazar\'{e}, Laurent and Grave, Edouard and D\'{e}fossez, Alexandre and Zeghidour, Neil},
  booktitle = 	 {Proceedings of the 42nd International Conference on Machine Learning},
  pages = 	 {32116--32129},
  year = 	 {2025},
  editor = 	 {Singh, Aarti and Fazel, Maryam and Hsu, Daniel and Lacoste-Julien, Simon and Berkenkamp, Felix and Maharaj, Tegan and Wagstaff, Kiri and Zhu, Jerry},
  volume = 	 {267},
  series = 	 {Proceedings of Machine Learning Research},
  month = 	 {13--19 Jul},
  publisher =    {PMLR},
  url = 	 {https://proceedings.mlr.press/v267/labiausse25a.html}
}

@misc{hibiki_zero_labiausse_2026,
      title={Simultaneous Speech-to-Speech Translation Without Aligned Data}, 
      author={Tom Labiausse and Romain Fabre and Yannick Estève and Alexandre Défossez and Neil Zeghidour},
      year={2026},
      eprint={2602.11072},
      archivePrefix={arXiv},
      primaryClass={cs.CL},
      url={https://arxiv.org/abs/2602.11072}, 
}

@misc{seamless_communication_2023,
  title={Seamless: Multilingual Expressive and Streaming Speech Translation},
  author={Barrault, Lo{\"\i}c and Chung, Yu-An and Meglioli, Mariano Coria and Dale, David and Dong, Ning and Duppenthaler, Mark and Duquenne, Paul-Ambroise and Ellis, Brian and Elsahar, Hady and Haaheim, Justin and others},
  journal={arXiv preprint arXiv:2312.05187},
  year={2023}
}

@misc{streamspeech_zhang_2024,
        title={StreamSpeech: Simultaneous Speech-to-Speech Translation with Multi-task Learning}, 
        author={Shaolei Zhang and Qingkai Fang and Shoutao Guo and Zhengrui Ma and Min Zhang and Yang Feng},
        year={2024},
        booktitle = {Proceedings of the 62th Annual Meeting of the Association for Computational Linguistics (Long Papers)},
        publisher = {Association for Computational Linguistics}
  }

@misc{seamlessM4T2023,
      title={SeamlessM4T: Massively Multilingual \& Multimodal Machine Translation},
      author={{Meta AI}},
      year={2023},
      eprint={2308.11596},
      archivePrefix={arXiv},
      primaryClass={cs.CL},
      url={https://arxiv.org/abs/2308.11596}, 
}

@inproceedings{federmann-etal-2022-ntrex,
    title = "{NTREX}-128 {--} News Test References for {MT} Evaluation of 128 Languages",
    author = "Federmann, Christian  and
      Kocmi, Tom  and
      Xin, Ying",
    editor = "Ahuja, Kabir  and
      Anastasopoulos, Antonios  and
      Patra, Barun  and
      Neubig, Graham  and
      Choudhury, Monojit  and
      Dandapat, Sandipan  and
      Sitaram, Sunayana  and
      Chaudhary, Vishrav",
    booktitle = "Proceedings of the First Workshop on Scaling Up Multilingual Evaluation",
    month = nov,
    year = "2022",
    address = "Online",
    publisher = "Association for Computational Linguistics",
    url = "https://aclanthology.org/2022.sumeval-1.4/",
    doi = "10.18653/v1/2022.sumeval-1.4",
    pages = "21--24"
}

@inproceedings{DPO2023,
 author = {Rafailov, Rafael and Sharma, Archit and Mitchell, Eric and Manning, Christopher D and Ermon, Stefano and Finn, Chelsea},
 booktitle = {Advances in Neural Information Processing Systems},
 editor = {A. Oh and T. Naumann and A. Globerson and K. Saenko and M. Hardt and S. Levine},
 pages = {53728--53741},
 publisher = {Curran Associates, Inc.},
 title = {Direct Preference Optimization: Your Language Model is Secretly a Reward Model},
 url = {https://proceedings.neurips.cc/paper_files/paper/2023/file/a85b405ed65c6477a4fe8302b5e06ce7-Paper-Conference.pdf},
 volume = {36},
 year = {2023}
}

@InProceedings{xu_cpo_2024,
  title = 	 {Contrastive Preference Optimization: Pushing the Boundaries of {LLM} Performance in Machine Translation},
  author =       {Xu, Haoran and Sharaf, Amr and Chen, Yunmo and Tan, Weiting and Shen, Lingfeng and Van Durme, Benjamin and Murray, Kenton and Kim, Young Jin},
  booktitle = 	 {Proceedings of the 41st International Conference on Machine Learning},
  pages = 	 {55204--55224},
  year = 	 {2024},
  editor = 	 {Salakhutdinov, Ruslan and Kolter, Zico and Heller, Katherine and Weller, Adrian and Oliver, Nuria and Scarlett, Jonathan and Berkenkamp, Felix},
  volume = 	 {235},
  series = 	 {Proceedings of Machine Learning Research},
  month = 	 {21--27 Jul},
  publisher =    {PMLR},
  url = 	 {https://proceedings.mlr.press/v235/xu24t.html}
}

@inproceedings{SimulPL2025,
 author = {Yu, Donglei and Zhao, Yang and Zhu, Jie and Xu, Yangyifan and Zhou, Yu and Zong, Chengqing},
 booktitle = {International Conference on Learning Representations},
 editor = {Y. Yue and A. Garg and N. Peng and F. Sha and R. Yu},
 pages = {55916--55938},
 title = {SimulPL: Aligning Human Preferences in Simultaneous Machine Translation},
 url = {https://proceedings.iclr.cc/paper_files/paper/2025/file/8c5111676c19e5945473acf20ad7026f-Paper-Conference.pdf},
 volume = {2025},
 year = {2025}
}

@inproceedings{seqpo_xu_2025,
    title = "{S}eq{PO}-{S}i{MT}: Sequential Policy Optimization for Simultaneous Machine Translation",
    author = "Xu, Ting  and
      Huang, Zhichao  and
      Sun, Jiankai  and
      Cheng, Shanbo  and
      Lam, Wai",
    editor = "Che, Wanxiang  and
      Nabende, Joyce  and
      Shutova, Ekaterina  and
      Pilehvar, Mohammad Taher",
    booktitle = "Findings of the Association for Computational Linguistics: ACL 2025",
    month = jul,
    year = "2025",
    address = "Vienna, Austria",
    publisher = "Association for Computational Linguistics",
    url = "https://aclanthology.org/2025.findings-acl.828/",
    doi = "10.18653/v1/2025.findings-acl.828",
    pages = "16107--16123",
    ISBN = "979-8-89176-256-5"
}

@inproceedings{berard2016,
  TITLE = {{Listen and Translate: A Proof of Concept for End-to-End Speech-to-Text Translation}},
  AUTHOR = {B{\'e}rard, Alexandre and Pietquin, Olivier and Besacier, Laurent and Servan, Christophe},
  URL = {https://hal.science/hal-01408086},
  BOOKTITLE = {{NIPS Workshop on end-to-end learning for speech and audio processing}},
  ADDRESS = {Barcelona, Spain},
  YEAR = {2016},
  MONTH = Dec,
  HAL_ID = {hal-01408086},
  HAL_VERSION = {v1},
}

@inproceedings{sacre_bleu_post_2018,
    title = "A Call for Clarity in Reporting {BLEU} Scores",
    author = "Post, Matt",
    editor = "Bojar, Ond{\v{r}}ej  and
      Chatterjee, Rajen  and
      Federmann, Christian  and
      Fishel, Mark  and
      Graham, Yvette  and
      Haddow, Barry  and
      Huck, Matthias  and
      Yepes, Antonio Jimeno  and
      Koehn, Philipp  and
      Monz, Christof  and
      Negri, Matteo  and
      N{\'e}v{\'e}ol, Aur{\'e}lie  and
      Neves, Mariana  and
      Post, Matt  and
      Specia, Lucia  and
      Turchi, Marco  and
      Verspoor, Karin",
    booktitle = "Proceedings of the Third Conference on Machine Translation: Research Papers",
    month = oct,
    year = "2018",
    address = "Brussels, Belgium",
    publisher = "Association for Computational Linguistics",
    url = "https://aclanthology.org/W18-6319/",
    doi = "10.18653/v1/W18-6319",
    pages = "186--191"
}

@misc{xcomet_guerreiro_2023,
  title={xcomet: Transparent machine translation evaluation through fine-grained error detection},
  author={Guerreiro, Nuno M and Rei, Ricardo and Stigt, Daan van and Coheur, Luisa and Colombo, Pierre and Martins, Andr{\'e} FT},
  journal={Transactions of the Association for Computational Linguistics},
  volume={12},
  pages={979--995},
  year={2024},
  publisher={MIT Press 255 Main Street, 9th Floor, Cambridge, Massachusetts 02142, USA~…}
}

@inproceedings{laal_papi_2022,
    title = "Over-Generation Cannot Be Rewarded: Length-Adaptive Average Lagging for Simultaneous Speech Translation",
    author = "Papi, Sara  and
      Gaido, Marco  and
      Negri, Matteo  and
      Turchi, Marco",
    editor = "Ive, Julia  and
      Zhang, Ruiqing",
    booktitle = "Proceedings of the Third Workshop on Automatic Simultaneous Translation",
    month = jul,
    year = "2022",
    address = "Online",
    publisher = "Association for Computational Linguistics",
    url = "https://aclanthology.org/2022.autosimtrans-1.2/",
    doi = "10.18653/v1/2022.autosimtrans-1.2",
    pages = "12--17"
}

@misc{silero_vad_2024,
  author = {Silero Team},
  title = {Silero VAD: pre-trained enterprise-grade Voice Activity Detector (VAD), Number Detector and Language Classifier},
  year = {2024},
  publisher = {GitHub},
  journal = {GitHub repository},
  howpublished = {\url{https://github.com/snakers4/silero-vad}},
  commit = {insert_some_commit_here},
  email = {hello@silero.ai}
}

@inproceedings{direct_simul_chen_2021,
    title = "Direct Simultaneous Speech-to-Text Translation Assisted by Synchronized Streaming {ASR}",
    author = "Chen, Junkun  and
      Ma, Mingbo  and
      Zheng, Renjie  and
      Huang, Liang",
    editor = "Zong, Chengqing  and
      Xia, Fei  and
      Li, Wenjie  and
      Navigli, Roberto",
    booktitle = "Findings of the Association for Computational Linguistics: ACL-IJCNLP 2021",
    month = aug,
    year = "2021",
    address = "Online",
    publisher = "Association for Computational Linguistics",
    url = "https://aclanthology.org/2021.findings-acl.406/",
    doi = "10.18653/v1/2021.findings-acl.406",
    pages = "4618--4624"
}

@article{Garcia2020,
    title={Taxing the bilingual mind: Effects of simultaneous interpreting experience on verbal and executive mechanisms},
    volume={23},
    DOI={10.1017/S1366728919000063},
    number={4},
    journal={Bilingualism: Language and Cognition},
    author={García, Adolfo M. and Muñoz, Edinson and Kogan, Boris},
    year={2020},
    pages={729–739}
}

@inproceedings{stacl_ma_2019,
    title = "{STACL}: Simultaneous Translation with Implicit Anticipation and Controllable Latency using Prefix-to-Prefix Framework",
    author = "Ma, Mingbo  and
      Huang, Liang  and
      Xiong, Hao  and
      Zheng, Renjie  and
      Liu, Kaibo  and
      Zheng, Baigong  and
      Zhang, Chuanqiang  and
      He, Zhongjun  and
      Liu, Hairong  and
      Li, Xing  and
      Wu, Hua  and
      Wang, Haifeng",
    editor = "Korhonen, Anna  and
      Traum, David  and
      M{\`a}rquez, Llu{\'i}s",
    booktitle = "Proceedings of the 57th Annual Meeting of the Association for Computational Linguistics",
    month = jul,
    year = "2019",
    address = "Florence, Italy",
    publisher = "Association for Computational Linguistics",
    url = "https://aclanthology.org/P19-1289/",
    doi = "10.18653/v1/P19-1289",
    pages = "3025--3036"
}

@inproceedings{christodoulides2014prosodic,
  title={Prosodic correlates of perceived quality and fluency in simultaneous interpreting},
  author={Christodoulides, George and Lenglet, C{\'e}dric},
  booktitle={Proceedings of the Speech Prosody},
  volume={7},
  pages={1002--1006},
  year={2014}
}

@inproceedings{whisperx_bain_2023,
  title     = {{WhisperX: Time-Accurate Speech Transcription of Long-Form Audio}},
  author    = {Max Bain and Jaesung Huh and Tengda Han and Andrew Zisserman},
  year      = {2023},
  booktitle = {{Interspeech 2023}},
  pages     = {4489--4493},
  doi       = {10.21437/Interspeech.2023-78},
  issn      = {2958-1796},
}

@inproceedings{common_voice_ardila_2020,
    title = "Common Voice: A Massively-Multilingual Speech Corpus",
    author = "Ardila, Rosana  and
      Branson, Megan  and
      Davis, Kelly  and
      Kohler, Michael  and
      Meyer, Josh  and
      Henretty, Michael  and
      Morais, Reuben  and
      Saunders, Lindsay  and
      Tyers, Francis  and
      Weber, Gregor",
    editor = "Calzolari, Nicoletta  and
      B{\'e}chet, Fr{\'e}d{\'e}ric  and
      Blache, Philippe  and
      Choukri, Khalid  and
      Cieri, Christopher  and
      Declerck, Thierry  and
      Goggi, Sara  and
      Isahara, Hitoshi  and
      Maegaard, Bente  and
      Mariani, Joseph  and
      Mazo, H{\'e}l{\`e}ne  and
      Moreno, Asuncion  and
      Odijk, Jan  and
      Piperidis, Stelios",
    booktitle = "Proceedings of the Twelfth Language Resources and Evaluation Conference",
    month = may,
    year = "2020",
    address = "Marseille, France",
    publisher = "European Language Resources Association",
    url = "https://aclanthology.org/2020.lrec-1.520/",
    pages = "4218--4222",
    language = "eng",
    ISBN = "979-10-95546-34-4"
}

@inproceedings{covost_wang_2021,
  title     = {{CoVoST 2 and Massively Multilingual Speech Translation}},
  author    = {Changhan Wang and Anne Wu and Jiatao Gu and Juan Pino},
  year      = {2021},
  booktitle = {{Interspeech 2021}},
  pages     = {2247--2251},
  doi       = {10.21437/Interspeech.2021-2027},
  issn      = {2958-1796},
}

@inproceedings{mturk_snow_2008,
    title = "Cheap and Fast {--} But is it Good? Evaluating Non-Expert Annotations for Natural Language Tasks",
    author = "Snow, Rion  and
      O{'}Connor, Brendan  and
      Jurafsky, Daniel  and
      Ng, Andrew",
    editor = "Lapata, Mirella  and
      Ng, Hwee Tou",
    booktitle = "Proceedings of the 2008 Conference on Empirical Methods in Natural Language Processing",
    month = oct,
    year = "2008",
    address = "Honolulu, Hawaii",
    publisher = "Association for Computational Linguistics",
    url = "https://aclanthology.org/D08-1027/",
    pages = "254--263"
}

@inproceedings{iwslt_ahmad-etal-2024-findings,
    title = "{FINDINGS} {OF} {THE} {IWSLT} 2024 {EVALUATION} {CAMPAIGN}",
    author = {Ahmad, Ibrahim Said  and
      Anastasopoulos, Antonios  and
      Bojar, Ond{\v{r}}ej  and
      Borg, Claudia  and
      Carpuat, Marine  and
      Cattoni, Roldano  and
      Cettolo, Mauro  and
      Chen, William  and
      Dong, Qianqian  and
      Federico, Marcello  and
      Haddow, Barry  and
      Javorsk{\'y}, D{\'a}vid  and
      Krubi{\'n}ski, Mateusz  and
      Lam, Tsz Kin  and
      Ma, Xutai  and
      Mathur, Prashant  and
      Matusov, Evgeny  and
      Maurya, Chandresh  and
      McCrae, John P.  and
      Murray, Kenton  and
      Nakamura, Satoshi  and
      Negri, Matteo  and
      Niehues, Jan  and
      Niu, Xing  and
      Ojha, Atul Kr.  and
      Ortega, John  and
      Papi, Sara  and
      Pol{\'a}k, Peter  and
      Posp{\'i}{\v{s}}il, Adam  and
      Pecina, Pavel  and
      Salesky, Elizabeth  and
      Sethiya, Nivedita  and
      Sarkar, Balaram  and
      Shi, Jiatong  and
      Sikasote, Claytone  and
      Sperber, Matthias  and
      St{\"u}ker, Sebastian  and
      Sudoh, Katsuhito  and
      Thompson, Brian  and
      Waibel, Alex  and
      Watanabe, Shinji  and
      Wilken, Patrick  and
      Zem{\'a}nek, Petr  and
      Zevallos, Rodolfo},
    editor = "Salesky, Elizabeth  and
      Federico, Marcello  and
      Carpuat, Marine",
    booktitle = "Proceedings of the 21st International Conference on Spoken Language Translation (IWSLT 2024)",
    month = aug,
    year = "2024",
    address = "Bangkok, Thailand (in-person and online)",
    publisher = "Association for Computational Linguistics",
    url = "https://aclanthology.org/2024.iwslt-1.1/",
    doi = "10.18653/v1/2024.iwslt-1.1",
    pages = "1--11"
}

\end{document}